\documentclass[10pt,twocolumn,letterpaper]{article}

\usepackage{dicta}
\usepackage{times}
\usepackage{epsfig}
\usepackage{graphicx}
\usepackage{amsmath}
\usepackage{amssymb}

\usepackage{makecell,adjustbox,booktabs}
\usepackage{graphicx}
\usepackage{booktabs}
\usepackage{tabularx}
\usepackage{multirow}
\usepackage{textcomp} 
\usepackage{xcolor}
\definecolor{lavender}{RGB}{179,157,219}     
\definecolor{coral}{RGB}{255,127,80}         
\definecolor{aquamarine}{RGB}{102,205,170}   
\definecolor{midgray}{RGB}{102,102,102}      

\usepackage{xcolor}

\usepackage[pagebackref=false,breaklinks=true,letterpaper=true,colorlinks,bookmarks=false]{hyperref}

\dictafinalcopy 

\def\dictaPaperID{85} 

\ifdictafinal\fi
\begin{document}

\title{Can Current AI Models Count What We Mean, Not What They See? \\ A Benchmark and Systematic Evaluation}

\author{%
  Gia Khanh Nguyen\textsuperscript{1}\quad
  Yifeng Huang\textsuperscript{2}\quad
  Minh Hoai\textsuperscript{1}\\[1ex]
  \textsuperscript{1}Australian Institute for Machine Learning, University of Adelaide, SA, Australia\\
  \textsuperscript{2}Stony Brook University, Stony Brook, NY, USA
}

\maketitle

\def\mA{\mathcal{A}}
\def\mB{\mathcal{B}}
\def\mC{\mathcal{C}}
\def\mD{\mathcal{D}}
\def\mE{\mathcal{E}}
\def\mF{\mathcal{F}}
\def\mG{\mathcal{G}}
\def\mH{\mathcal{H}}
\def\mI{\mathcal{I}}
\def\mJ{\mathcal{J}}
\def\mK{\mathcal{K}}
\def\mL{\mathcal{L}}
\def\mM{\mathcal{M}}
\def\mN{\mathcal{N}}
\def\mO{\mathcal{O}}
\def\mP{\mathcal{P}}
\def\mQ{\mathcal{Q}}
\def\mR{\mathcal{R}}
\def\mS{\mathcal{S}}
\def\mT{\mathcal{T}}
\def\mU{\mathcal{U}}
\def\mV{\mathcal{V}}
\def\mW{\mathcal{W}}
\def\mX{\mathcal{X}}
\def\mY{\mathcal{Y}}
\def\mZ{\mathcal{Z}} 

\def\bbN{\mathbb{N}} 
\def\bbR{\mathbb{R}} 
\def\bbP{\mathbb{P}} 
\def\bbQ{\mathbb{Q}} 
\def\bbE{\mathbb{E}}

\def\1n{\mathbf{1}_n}
\def\0{\mathbf{0}}
\def\1{\mathbf{1}}

\def\A{{\bf A}}
\def\B{{\bf B}}
\def\C{{\bf C}}
\def\D{{\bf D}}
\def\E{{\bf E}}
\def\F{{\bf F}}
\def\G{{\bf G}}
\def\H{{\bf H}}
\def\I{{\bf I}}
\def\J{{\bf J}}
\def\K{{\bf K}}
\def\L{{\bf L}}
\def\M{{\bf M}}
\def\N{{\bf N}}
\def\O{{\bf O}}
\def\P{{\bf P}}
\def\Q{{\bf Q}}
\def\R{{\bf R}}
\def\S{{\bf S}}
\def\T{{\bf T}}
\def\U{{\bf U}}
\def\V{{\bf V}}
\def\W{{\bf W}}
\def\X{{\bf X}}
\def\Y{{\bf Y}}
\def\Z{{\bf Z}}

\def\a{{\bf a}}
\def\b{{\bf b}}
\def\c{{\bf c}}
\def\d{{\bf d}}
\def\e{{\bf e}}
\def\f{{\bf f}}
\def\g{{\bf g}}
\def\h{{\bf h}}
\def\i{{\bf i}}
\def\j{{\bf j}}
\def\k{{\bf k}}
\def\l{{\bf l}}
\def\m{{\bf m}}
\def\n{{\bf n}}
\def\o{{\bf o}}
\def\p{{\bf p}}
\def\q{{\bf q}}
\def\r{{\bf r}}
\def\s{{\bf s}}
\def\t{{\bf t}}
\def\u{{\bf u}}
\def\v{{\bf v}}
\def\w{{\bf w}}
\def\x{{\bf x}}
\def\y{{\bf y}}
\def\z{{\bf z}}

\def\balpha{\mbox{\boldmath{$\alpha$}}}
\def\bbeta{\mbox{\boldmath{$\beta$}}}
\def\bdelta{\mbox{\boldmath{$\delta$}}}
\def\bgamma{\mbox{\boldmath{$\gamma$}}}
\def\blambda{\mbox{\boldmath{$\lambda$}}}
\def\bsigma{\mbox{\boldmath{$\sigma$}}}
\def\btheta{\mbox{\boldmath{$\theta$}}}
\def\bomega{\mbox{\boldmath{$\omega$}}}
\def\bxi{\mbox{\boldmath{$\xi$}}}
\def\bnu{\mbox{\boldmath{$\nu$}}}                                  
\def\bphi{\mbox{\boldmath{$\phi$}}}
\def\bmu{\mbox{\boldmath{$\mu$}}}

\def\bDelta{\mbox{\boldmath{$\Delta$}}}
\def\bOmega{\mbox{\boldmath{$\Omega$}}}
\def\bPhi{\mbox{\boldmath{$\Phi$}}}
\def\bLambda{\mbox{\boldmath{$\Lambda$}}}
\def\bSigma{\mbox{\boldmath{$\Sigma$}}}
\def\bGamma{\mbox{\boldmath{$\Gamma$}}}
                                  
\newcommand{\myprob}[1]{\mathop{\mathbb{P}}_{#1}}

\newcommand{\myexp}[1]{\mathop{\mathbb{E}}_{#1}}

\newcommand{\mydelta}[1]{1_{#1}}

\newcommand{\myminimum}[1]{\mathop{\textrm{minimum}}_{#1}}
\newcommand{\mymaximum}[1]{\mathop{\textrm{maximum}}_{#1}}    
\newcommand{\mymin}[1]{\mathop{\textrm{minimize}}_{#1}}
\newcommand{\mymax}[1]{\mathop{\textrm{maximize}}_{#1}}
\newcommand{\mymins}[1]{\mathop{\textrm{min.}}_{#1}}
\newcommand{\mymaxs}[1]{\mathop{\textrm{max.}}_{#1}}  
\newcommand{\myargmin}[1]{\mathop{\textrm{argmin}}_{#1}} 
\newcommand{\myargmax}[1]{\mathop{\textrm{argmax}}_{#1}} 
\newcommand{\myst}{\textrm{s.t. }}

\newcommand{\denselist}{\itemsep -1pt}
\newcommand{\sparselist}{\itemsep 1pt}

\definecolor{pink}{rgb}{0.9,0.5,0.5}
\definecolor{purple}{rgb}{0.5, 0.4, 0.8}   
\definecolor{gray}{rgb}{0.3, 0.3, 0.3}
\definecolor{mygreen}{rgb}{0.2, 0.6, 0.2}

\newcommand{\cyan}[1]{\textcolor{cyan}{#1}}
\newcommand{\red}[1]{\textcolor{red}{#1}}  
\newcommand{\blue}[1]{\textcolor{blue}{#1}}
\newcommand{\magenta}[1]{\textcolor{magenta}{#1}}
\newcommand{\pink}[1]{\textcolor{pink}{#1}}
\newcommand{\green}[1]{\textcolor{green}{#1}} 
\newcommand{\gray}[1]{\textcolor{gray}{#1}}    
\newcommand{\mygreen}[1]{\textcolor{mygreen}{#1}}    
\newcommand{\purple}[1]{\textcolor{purple}{#1}}       

\definecolor{greena}{rgb}{0.4, 0.5, 0.1}
\newcommand{\greena}[1]{\textcolor{greena}{#1}}

\definecolor{bluea}{rgb}{0, 0.4, 0.6}
\newcommand{\bluea}[1]{\textcolor{bluea}{#1}}
\definecolor{reda}{rgb}{0.6, 0.2, 0.1}
\newcommand{\reda}[1]{\textcolor{reda}{#1}}

\def\changemargin#1#2{\list{}{\rightmargin#2\leftmargin#1}\item[]}
\let\endchangemargin=\endlist
                                               
\newcommand{\cm}[1]{}

\newcommand{\mhoai}[1]{{\color{magenta}\textbf{[MH: #1]}}}
\newcommand{\yifeng}[1]{{\color{blue}\textbf{[yifeng: #1]}}}

\newcommand{\mtodo}[1]{{\color{red}$\blacksquare$\textbf{[TODO: #1]}}}
\newcommand{\myheading}[1]{\vspace{0.5ex}\noindent \textbf{#1}}
\newcommand{\htimesw}[2]{\mbox{$#1$$\times$$#2$}}


%
%
%

\newcommand{\Sref}[1]{Sec.~\ref{#1}}
\newcommand{\Eref}[1]{Eq.~(\ref{#1})}
\newcommand{\Fref}[1]{Fig.~\ref{#1}}
\newcommand{\Tref}[1]{Table~\ref{#1}}

\begin{abstract}
Visual counting is a fundamental yet challenging task, especially when users need to count objects of a specific type in complex scenes. While recent models, including class-agnostic counting models and large vision-language models (VLMs), show promise in counting tasks, their ability to perform fine-grained, intent-driven counting remains unclear. In this paper, we introduce PairTally, a benchmark dataset specifically designed to evaluate fine-grained visual counting. Each of the 681 high-resolution images in PairTally contains two object categories, requiring models to distinguish and count based on subtle differences in shape, size, color, or semantics. The dataset includes both inter-category (distinct categories) and intra-category (closely related subcategories) settings, making it suitable for rigorous evaluation of selective counting capabilities. We benchmark a variety of state-of-the-art models, including exemplar-based methods, language-prompted models, and large VLMs. Our results show that despite recent advances, current models struggle to reliably count what users intend, especially in fine-grained and visually ambiguous cases. PairTally provides a new foundation for diagnosing and improving fine-grained visual counting systems. The data and code for this benchmark are available at \url{https://github.com/bbvisual/PairTally_Benchmark}.


\end{abstract}

\section{Introduction}
\begin{figure*}[t]
  \centering
  \includegraphics[width=\textwidth,page=1]{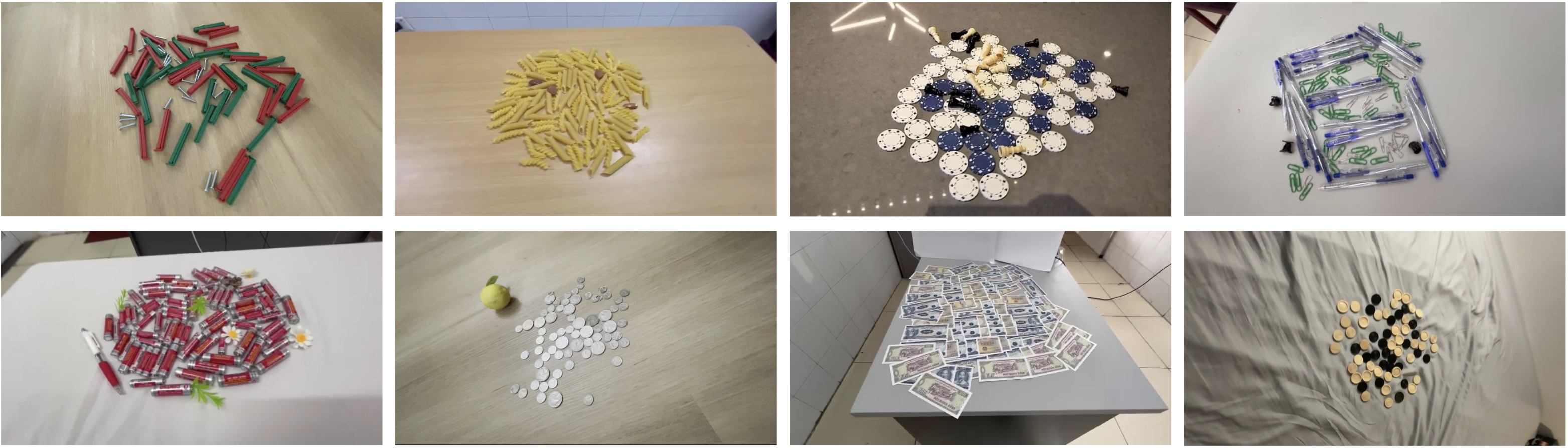}
  \caption{Example samples from the PairTally dataset. Each panel shows intra‐ and inter‐category scenes with fine‐grained object variants. \label{fig:dataset_sample}}
  
\end{figure*}

Visual counting is needed in a wide range of situations, from tallying boxes for inventory management and counting cells for disease diagnostics. With the advent of large vision-language models (VLMs)—capable of detecting and recognizing objects and interpreting complex visual scenes—it is natural to explore whether such tasks can now be delegated to machines, particularly in scenarios where multiple object categories are present and models must infer what the user intends to count.

To answer that question, we must first consider how human users might communicate their counting intent to AI models. This is typically done through categorical naming, spatial delineation, or language-based reference. While some of these forms, such as exemplar-based bounding boxes, have been supported by traditional visual counting methods developed before the rise of VLMs, others, particularly natural language instructions and multi-modal integration, are only feasible with more recent advances. Earlier approaches included class-specific models~\cite{li2018csrnet,miao2020shallow,jiang2020attention,wang2020DMCount,xu2019learn,hu2020count,bai2020adaptive,liu2020adaptive,song2021rethinking,liu2019point,lian2019density,sam2020locate,liu2019recurrent,laradji2018blobs,m_Ranjan-etal-ECCV18,m_Ranjan-etal-ACCV20,m_Abousamra-etal-AAAI21} designed for particular object categories such as people, cars, or cells. Later, class-agnostic methods~\cite{lu2018class, ranjan2021learning, yang2021class,  shi2022represent, nguyen2022few, m_Ranjan-Hoai-ACCV22, m_Huang-etal-ICCV23} have been developed to count arbitrary objects of interest, given a few exemplars delineated by bounding boxes. The advent of foundation models and large VLMs has significantly enhanced communication flexibility, enabling richer ways to specify counting targets through natural language, visual cues, or a combination of both. Modern counting models~\cite{amini2024countgd, Dai_2024_CVPR} can now process an input image along with diverse prompts to identify the objects to be counted, returning a predicted total. VLMs can even support multi-turn conversations, allowing users to provide feedback and request a recount.



While communication with AI models has become more flexible, it is unclear how much their counting accuracy has improved. Do models truly attend to the user’s specified target, or do they default to counting the most visually dominant categories, as reinforced by existing datasets and benchmarks? Moreover, can they distinguish targets based on fine-grained attributes such as shape, size, color, or subtle appearance differences?


In this paper, we benchmark ten state-of-the-art visual counting models, organized by their prompting mechanisms and underlying architectures. Our selection comprises four class-agnostic counters—FamNet~\cite{ranjan2021learning}, DAVE~\cite{pelhan2024dave}, GeCo~\cite{pelhan2024novel}, and LoCA~\cite{djukic2023loca}, which regress density maps from bounding-box exemplars (although DAVE support text prompts through CLIP, we use DAVE exclusively in its bounding-box mode, where it performs best). Next are two object detectors: CountGD~\cite{amini2024countgd}, built on Grounding DINO and fine-tuned for counting with both text and bounding-box prompts, and LLMDet~\cite{fu2025llmdet}, a text-only object detector not trained specifically for counting. Finally, we evaluate four large vision–language models, Ovis2~\cite{lu2024ovis}, Qwen2.5‑VL~\cite{Qwen2.5-VL}, InternVL3~\cite{zhu2025internvl3exploringadvancedtraining}, and LLaMA‑3.2~\cite{dubey2024llama}, which perform counting via multimodal instruction following. Together, these ten methods span modern counting paradigms, from dense regression to flexible, promptable detectors and instruction-tuned foundation models.


To evaluate these models thoroughly, we need a dataset in which each image contains a substantial number of objects from at least one category, justifying the need for counting, while also including other objects that could plausibly be mistaken for the target category, as often occurs in real-world settings. Unfortunately, existing datasets either feature low object counts across all categories or include many objects predominantly from a single category. Such datasets are inadequate for evaluating fine-grained counting under realistic, challenging conditions. 

To address this gap, we introduce {\bf PairTally}, a diagnostic dataset for fine-grained counting. PairTally contains 681 high-resolution images, each featuring two object categories with substantial instance counts (\Fref{fig:dataset_sample}). About half of the pairs involve clearly distinct categories, while the rest contrast subtle subcategories varying in shape, size, or color. Scenes are drawn mainly from tabletop and household contexts for controlled and repeatable setups. The aim is not exhaustive coverage, but targeted analysis of failure modes likely to arise in broader, open-world scenarios.

We use PairTally to examine key challenges in fine-grained visual counting:  
(1) Do models follow counting prompts reliably?  
(2) Can they distinguish two object categories in the same scene, whether distinct or subtle variants of the same class?  
(3) How sensitive are they to fine-grained attributes such as color, size, and texture/shape?  
This study provides the first systematic evaluation of these challenges in open-world fine-grained counting.


Overall, two consistent patterns emerge. Specialist models tend to overcount, drifting toward counting everything because they indiscriminately enumerate repeated patterns. In contrast, VLMs tend to undercount, as they were not trained for counting but rather for general detection and recognition tasks. Together, these biases show that current methods fall short in fine-grained counting, necessitating models that can both count accurately and distinguish visually similar objects.

\section{Related Work}



\myheading{Benchmark Datasets.} Early progress in object counting has been shaped by domain-specific datasets such as CARPK~\cite{hsieh2017drone}, NDISPark~\cite{visapp21}, TRANCOS~\cite{TRANCOSdataset_IbPRIA2015}, ShanghaiTech~\cite{zhang2016single}, JHU-CROWD++~\cite{sindagi2020jhu}, and VGG-Cell~\cite{xie2018microscopy}. While effective within narrow applications—vehicles, crowds, or cells—these datasets are unsuitable for evaluating general-purpose counting in open-world, multi-category scenes. FSC-147~\cite{ranjan2021learning} and FSCD-147~\cite{nguyen2022few} introduced broader category diversity and remains the dominant benchmark for class-agnostic counting. However, each image of this dataset contains mostly object instances from a single object class, leaving multiclass and fine-grained counting effectively untested.

Recent datasets OmniCount‑191~\cite{mondal2025omnicount}, CountBench~\cite{paiss2023teaching}, FSCD-LVIS~\cite{nguyen2022few}, and MCAC~\cite{hobley2024abc} push beyond FSC‑147 by introducing greater category diversity, bounding box annotations, and multi‑object scenes. However, none offer {controlled, real‑world} images that (i) require counting two object types simultaneously and (ii) demand discrimination between visually similar subtypes. While OmniCount‑191 supports multiclass annotations, its coarse grouping and unstructured image compositions fall short of the systematic two‑object tests that PairTally delivers. 


\myheading{Benchmark studies.} 
While the literature contains surveys of class‑agnostic counting methods, most comprehensively Ciampi\etal \cite{ciampi2025survey}, there is surprisingly little empirical work that systematically dissects what current models can and cannot count. The recent PrACo benchmark~\cite{ciampi2024praco} makes a valuable contribution by introducing tests for prompt sensitivity. However, despite this advance, PrACo falls short of testing a model's ability to perform fine-grained and multi-class counting under realistic conditions. Its mosaic test images are synthetically composed by stitching together crops of single-category scenes, meaning models are never required to interpret two object types in the same physical context or deal with natural clutter, occlusion, or layout. As mentioned above due to a lack of multi-class datasets, there is also a lack of empirical studies that assess fine-grained counting abilities. Our study is the first systematic, cross‑paradigm audit under tests that simultaneously demand (i) distinguishing different object types and (ii) discriminating subtle intra‑category differences (color, size, shape); our analysis thus fills a critical gap between broad surveys and ad‑hoc model demonstrations.

\section{State of the art counting models}



\myheading{Counting models that use visual exemplars.}
We benchmark four counting models from this category: FamNet~\cite{ranjan2021learning}, LOCA~\cite{djukic2023loca}, GeCo~\cite{pelhan2024novel}, and DAVE~\cite{pelhan2024dave}. These follow the exemplar-guided paradigm, where one or more visual exemplars are provided at test time, and the model must count similar objects in a query image. 

We do not benchmark several other exemplar-based methods. CACViT~\cite{wang2024vision}, ConCoNet~\cite{soliven2023conconet}, SAFECount~\cite{you2023few}, and BMNet+~\cite{shi2022represent} either do not achieve state-of-the-art results or offer limited novelty over baselines being tested. Our selected models are representative, influential, and well-suited to the exemplar-guided setting we evaluate.

\myheading{Counting models that use language prompts}
Prompt-based and open-world counting methods aim to generalize beyond fixed category sets, typically without relying on explicit visual exemplars. These models leverage natural language prompts to guide object localization and counting, enabling flexible and open-ended reasoning.

We benchmark CountGD and LLMDet, which represent a shift from the exemplar-based approach to the prompt-compatible counting approach. Both use open-vocabulary detectors to predict bounding boxes based on textual descriptions, bypassing the need for reference crops. CountGD builds on GroundingDINO~\cite{liu2024grounding}, while LLMDet~\cite{fu2025llmdet} is a newer object detection model that achieves improved accuracy and grounding precision via language-guided detection and counting.

Other text-based methods such as CLIP-Count~\cite{jiang2023clip}, VLCounter~\cite{kang2024vlcounter}, CounTX~\cite{amini2023open}, ZSC~\cite{xu2023zero}, VA-Count~\cite{zhu2024zero}, PseCo~\cite{huang2024point} and TFPOC~\cite{lin2024fixed} are not included due to limited performance on recent benchmarks. GroundingREC~\cite{Dai_2024_CVPR} is excluded as its referring expression comprehension design does not transfer well to our dataset format.

\myheading{Counting with Vision-Language Models.}
Recent VLMs such as Ovis2~\cite{lu2024ovis}, Qwen2.5-VL~\cite{Qwen2.5-VL}, LLaMA-3.2~\cite{dubey2024llama}, and InternVL3~\cite{zhu2025internvl3exploringadvancedtraining} represent a new generation of multimodal systems that combine large language models with visual encoders. These models are commonly used for tasks such as image captioning, visual question answering, and referring expression comprehension, where they generate text responses grounded in visual input. To evaluate their capabilities in fine-grained visual counting, we selected mid- to smaller-sized variants---Ovis2 (16B), Qwen2.5-VL (7B), LLaMA-3.2 (11B), and InternVL3 (14B)---rather than their largest counterparts. This decision was driven by practical considerations around resource constraints, and accessibility. By focusing on more deployable model sizes, we aim to provide insights that are both scalable and relevant to real-world use cases, while still capturing the core strengths and limitations of these architectures.


We do not benchmark commercial models (e.g., ChatGPT, Gemini) due to their proprietary nature, which hinders reproducibility. Our findings therefore focus on open-source VLMs. While we make no formal claims about commercial models, we believe they are unlikely to substantially outperform those tested, based on known performance gaps and limited preliminary experiments.


\section{PairTally -- a New Benchmark Dataset}



\begin{table}[t]
\centering
\scriptsize
\renewcommand{\arraystretch}{1.1}
\begin{tabularx}{\linewidth}{@{}lX@{}}
\toprule
\textbf{Supercategory} & \textbf{Category (Subcategory 1, Subcategory 2)} \\
\midrule
\textbf{Food} & \parbox[t]{\linewidth}{%
\textcolor{aquamarine}{pasta (spiral, penne)}, 
\textcolor{coral}{lime (citrus, calamansi)}, 
\textcolor{lavender}{peppercorn (black, white)}, 
\textcolor{coral}{tomato (normal, baby)}, 
\textcolor{aquamarine}{chili (long, short)}, 
\textcolor{lavender}{peanut (with/without skin)}, 
\textcolor{lavender}{bean (black, soy)}, 
\textcolor{aquamarine}{seed (pumpkin, sunflower)}, 
\textcolor{lavender}{coffee candy (brown, black)}, 
\textcolor{midgray}{garlic}, 
\textcolor{midgray}{shallot}} \\
\midrule
\textbf{Fun} & \parbox[t]{\linewidth}{%
\textcolor{lavender}{checker piece (black, white)}, 
\textcolor{aquamarine}{mahjong tile (bamboo, character)}, 
\textcolor{lavender}{lego piece (green, pink)}, 
\textcolor{lavender}{chess piece (black, white)}, 
\textcolor{aquamarine}{puzzle piece (edge, center)}, 
\textcolor{aquamarine}{puzzle piece (edge, center)}, 
\textcolor{lavender}{poker chip (blue, white)}, 
\textcolor{lavender}{playing card (red, black)}, 
\textcolor{coral}{marble (big, small)}, 
\textcolor{lavender}{dice (green, white)}, 
\textcolor{aquamarine}{Chinese card (red, black)}} \\
\midrule
\textbf{Household} & \parbox[t]{\linewidth}{%
\textcolor{aquamarine}{toothpick (straight, plastic)}, 
\textcolor{aquamarine}{cotton bud (wooden, plastic)}, 
\textcolor{lavender}{pill (white, yellow)}, 
\textcolor{coral}{battery (AAA, AA)}, 
\textcolor{lavender}{hair clipper (black, brown)}, 
\textcolor{lavender}{bill (1000, 5000 VND)}, 
\textcolor{coral}{coin (5¢, 10¢)}, 
\textcolor{aquamarine}{bottle cap (beer, plastic)}, 
\textcolor{aquamarine}{shirt button (2, 4 holes)}, 
\textcolor{aquamarine}{utensil (spoon, fork)}} \\
\midrule
\textbf{Office} & \parbox[t]{\linewidth}{%
\textcolor{aquamarine}{push pin (normal, round)}, 
\textcolor{coral}{heart sticker (big, small)}, 
\textcolor{lavender}{craft stick (red/orange, blue/purple)}, 
\textcolor{lavender}{rubber band (yellow, blue)}, 
\textcolor{lavender}{sticky note (green shades)}, 
\textcolor{lavender}{paper clip (silver, colored)}, 
\textcolor{aquamarine}{pen (with/without cap)}, 
\textcolor{midgray}{pencil}, 
\textcolor{aquamarine}{rhinestone (round, star)}, 
\textcolor{coral}{zip tie (short, long)}, 
\textcolor{coral}{safety pin (big, small)}, 
\textcolor{midgray}{lapel pin}, 
\textcolor{midgray}{wire organizer}} \\
\midrule
\textbf{Other} & \parbox[t]{\linewidth}{%
\textcolor{aquamarine}{screw (bronze, silver)}, 
\textcolor{aquamarine}{bolt (hex, mushroom)}, 
\textcolor{aquamarine}{nut (hex, square)}, 
\textcolor{aquamarine}{washer (metal, nylon)}, 
\textcolor{lavender}{bead (blue/pink shades)}, 
\textcolor{lavender}{ikea clip (green, red)}, 
\textcolor{lavender}{peg (grey, white)}, 
\textcolor{lavender}{stone (red, yellow)}, 
\textcolor{lavender}{novelty buttons (beige, transparent)}} \\
\bottomrule
\end{tabularx}
\vskip 0.05in
\caption{Object categories of the PairTally dataset. Subcategory division are based on:
\textcolor{lavender}{color}, \textcolor{coral}{size}, \textcolor{aquamarine}{texture/shape}, \textcolor{midgray}{no subcategories}. \label{tab:pairtally_folds}}

\end{table}


\myheading{Pair Construction and Scene Design}. PairTally is organized into five broad \emph{supercategories}: Food, Fun, Household, Office, and Other. The dataset comprises 54 object \emph{categories}. For each category (e.g.\ \emph{poker chips}, \emph{pens}), we manually defined up to two visually distinct \emph{subcategories} based on attributes of interest: color (43.5\% of subcategory distinctions), texture or shape (42.5\%), and size (14.1\%). This was done in a way that ensure real-world relevance simulating scenarios that a human would go through such as counting game pieces or sorting office items. We prioritized these real life examples to capture the kinds of visual variation that would occur in daily life. This yielded 98 subcategories in total as shown in \Tref{tab:pairtally_folds}.


We next systematically constructed an initial set of 100 \emph{subcategory pairs}: 50 \textbf{intra‑category} pairs that match the two subcategories within each category, and 50 \textbf{inter‑category} pairs chosen across categories. Later on, we excluded three intra-category pairs due to insufficient visual distinction (e.g., ``light green vs.\ mint green cups''). Ambiguities also emerged when each subclass contained more than three fine-grained variations, yet some models allowed only three exemplar bounding boxes per query. For instance, in ``small vs.\ big M\&Ms,'' both sizes appeared in various colors (red, yellow, green, blue), making it unclear whether the model should count all items of a given size or only those matching the color of the exemplars. To ensure consistent interpretation and fair evaluation, we excluded such ambiguous cases from the benchmark. The final set comprises 97 subcategory pairs.

\myheading{Data capture and annotation.}
All scenes were recorded using consumer-grade smartphones (iPhone 12, Samsung S21 Plus), capturing natural diversity in lighting, perspective, occlusion, and scene complexity. Each image was annotated with the following: \textbf{three bounding boxes} each (for exemplar based counting), \textbf{textual labels} identifying object categories and fine-grained variants (for prompt based counting), \textbf{count labels} denoting the number of instances per object type in each scene. 

\section{Experiments}
For all evaluations, each model was asked to count the same things, differing only in how exemplars and text prompts were supplied. DAVE, GeCo, LoCA, and FamNet received three exemplar bounding boxes per scene; CountGD was evaluated in two modes—exemplar\,+\,text (three exemplar boxes and a textual reference) and text‑only; LLMDet operated in text‑only mode (textual reference); and Vision–Language Models (Ovis2, Qwen2.5‑VL, LLaMA‑3.2, InternVL3) were queried with the unified prompt:  
``Count the number of \{object\} in this image. Provide only the total count in this format: \textless count\textgreater N\textless /count\textgreater. If you see no \{object\} or are unsure, respond with \textless count\textgreater 0\textless /count\textgreater.''  
The placeholder \{object\}'' was replaced with the specific reference (e.g., red poker chips''), and counts were extracted by parsing the integer within the \textless count\textgreater…\textless /count\textgreater\ tags.  

\subsection{Main results}

Recall that each image in our benchmark dataset contains objects from two categories, along with a few possible distractor objects from other classes. The two main categories are unordered, but for convenience, we refer to them as $A$ and $B$, with $a$ and $b$ denoting the number of objects in each category, respectively. For a given method, let $f(A)$ represent the predicted count when the model is prompted to count objects in category $A$, and let $f(A + B)$ denote the predicted count when the model is asked to count both $A$ and $B$. In the latter case, if the method requires bounding-box exemplars, we provide two from $A$ and one from $B$.

To evaluate the accuracy of each method, we use the absolute difference between the predicted count and the ground truth. Specifically, we compute $|f(A) - a|$ and $|f(A + B) - (a + b)|$ to assess how well the predicted counts match the expected values for each prompt.

To further analyze model behavior, we also compute $|f(A) - (a + b)|$ and compare it with $|f(A) - a|$ to evaluate whether the model mistakenly counts objects from category $B$ when only $A$ is requested. Additionally, we compute $|f(A) - f(A + B)|$ to assess whether the model is responsive to the prompt---that is, whether it adjusts its predictions based on the requested categories or tends to produce similar outputs regardless of the prompt content.

\begin{figure}[htbp]
\centering
\includegraphics[width=\columnwidth]{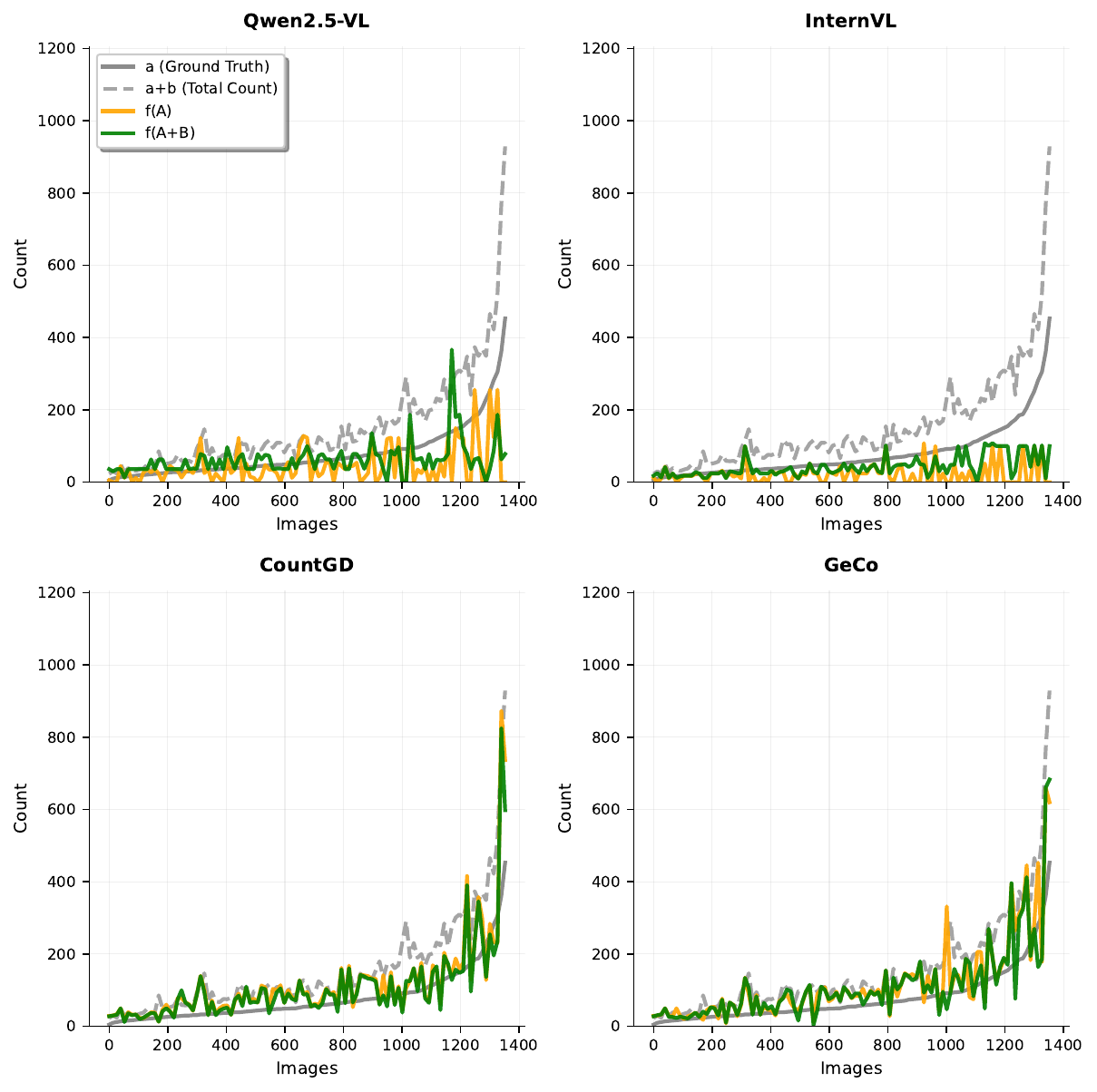}
\caption{Model counting predictions sorted by increasing ground truth count. Solid grey: ground truth queried category a. Dashed grey: total count a+b. Orange: predicted count f(A) when querying single category. Green: predicted count f(A+B) when querying both categories.}
\label{fig:counting_comparison}
\end{figure}

\begin{figure*}[t]
    \centering
    \includegraphics[width=0.8\linewidth]{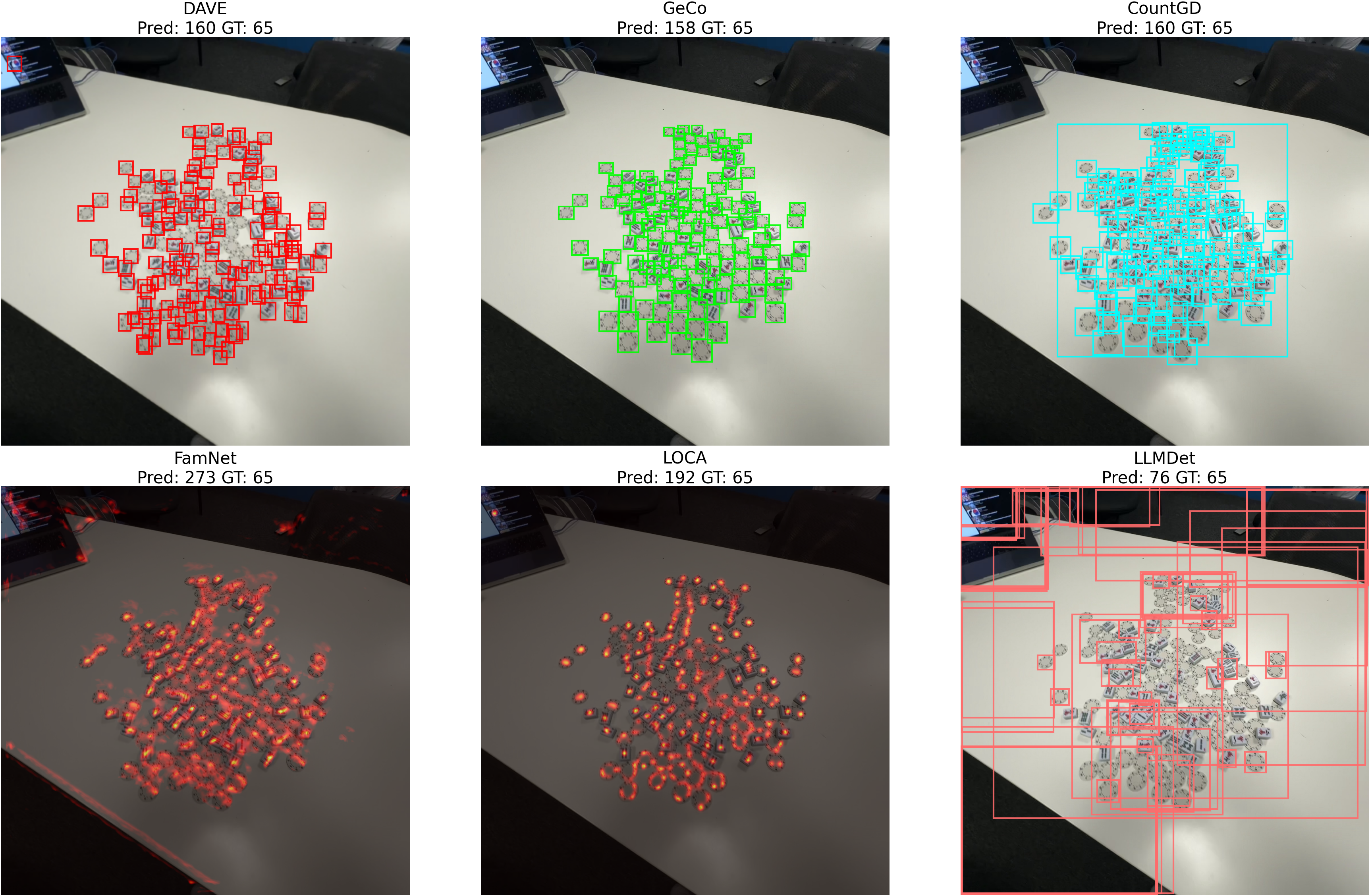}
    \caption{Inter-category scene with Mahjong tiles and poker chips; models were asked to count only Mahjong tiles.
    \label{fig:qualitative_grid_inter}
    }
    
\end{figure*}

\begin{table*}[t]
\centering
\setlength{\tabcolsep}{4pt} 
\begin{adjustbox}{max width=\textwidth}
\begin{tabular}{llcccccc}
\toprule
& & \multicolumn{4}{c}{Mean over test images} & \multicolumn{2}{c}{Percentage over test images} \\
\cmidrule(lr){3-6} \cmidrule(lr){7-8}
Model & Prompt &
\makecell{$|f(A)-a|$} &
\makecell{$|f(A+B)-(a+b)|$} &
\makecell{$|f(A)-(a+b)|$} &
\makecell{$|f(A)-f(A+B)|$} &
\makecell{$f(A) > a$} &
\makecell{$|f(A)-a| > |f(A)-(a+b)|$} \\
\midrule
DAVE \cite{pelhan2024dave} & b.boxes & 47.37 & \ 69.49 & \ 69.89 & 18.00 & 74.0 & 48.6 \\
GeCo \cite{pelhan2024novel} & b.boxes & 50.24 & \ 53.07 & \ 51.81 & 21.17 & 84.7 & 59.8 \\
LOCA \cite{djukic2023loca} & b.boxes & 72.80 & \ 62.78 & \ 76.51 & 39.73 & 83.1 & 58.6 \\
FamNet \cite{ranjan2021learning} & b.boxes & 75.83 & \ 88.30 & \ 90.60 & 51.57 & 73.1 & 55.7 \\
\midrule
Count GD \cite{amini2024countgd} & both & 46.67 & \ 57.33 & \ 52.88 & 10.58 & 83.1 & 57.3 \\
Count GD (Text) \cite{amini2024countgd} & text & 50.32 & \ 57.33 & \ 93.99 & 42.83 & 52.0 & 32.9 \\
LLMDet \cite{fu2025llmdet} & text & 77.23 & 107.84 & 121.82 & 45.49 & 53.3 & 41.4 \\
\midrule
Ovis2 \cite{lu2024ovis} & text & 65.15 & 111.56 & 145.37 & 35.69 & \ 5.0 & \ 0.3 \\
Qwen2.5-VL \cite{Qwen2.5-VL} & text & 59.36 & \ 99.88 & 126.88 & 43.62 & 26.1 & 10.7 \\
LLaMA-3.2 \cite{dubey2024llama} & text & 54.67 & \ 97.56 & 130.46 & 45.15 & 12.9 & \ 2.5 \\
InternVL3 \cite{zhu2025internvl3exploringadvancedtraining} & text & 63.41 & 115.98 & 142.10 & 27.59 & \ 9.1 & \ 1.0 \\
\bottomrule
\end{tabular}
\end{adjustbox}
\vskip 0.05in
\caption{Fine-grained counting results comparing single-category $f(A)$ with dual-category $f(A+B)$ against ground truths $a$ and $b$.}

\label{tab:fine-grained-counting}
\end{table*}

The results in Figure~\ref{fig:counting_comparison} and Table~\ref{tab:fine-grained-counting} reveal distinct failure modes across model categories. Vision-Language Models (e.g., Qwen2.5-VL, InternVL) exhibit erratic and inconsistent predictions that often bear little correlation to either the ground truth or total counts. Their outputs fluctuate unpredictably across images, suggesting fundamental difficulties in the visual counting task itself, with predictions largely disconnected from the true object quantities.

By contrast, specialist counting models (e.g., CountGD, GeCo) display a more systematic but equally problematic behavior: overcounting. When prompted with a single category $A$, these models frequently predict $f(A) > a$, as shown in Table~\ref{tab:fine-grained-counting}, where over 70\% of cases for DAVE and FamNet and higher rates for GeCo and CountGD exceed the true count. Moreover, for GeCo and CountGD, $|f(A)-a| > |f(A)-(a+b)|$ more than half of the time, indicating that their predictions drift toward the total object count rather than the queried subset. In other words, these models perceive object quantities but systematically conflate $A$ and $B$, counting both even when only $A$ is requested.

Figure~\ref{fig:counting_comparison} illustrates this contrast. The orange curves ($f(A)$) of specialist models often track the dashed line ($a+b$) more closely than the solid line ($a$), a clear sign of category confusion and overcounting. Conversely, their green curves ($f(A+B)$) align more closely with $a+b$, showing that they are more reliable when tasked with total counts rather than fine-grained ones. VLMs, on the other hand, show no such consistent structure, underscoring their weakness in the basic counting task itself.

Overall, these results suggest that while VLMs fail at robust counting altogether, specialist models succeed at estimating totals but struggle to follow prompts for fine-grained counting. This indicates that their limitation lies not in visual perception but in instruction-following, whereas VLMs face the challenge of weak visual counting overall. The contrasting failure modes point to fundamentally different architectural bottlenecks that must be addressed for reliable fine-grained counting.


\Fref{fig:qualitative_grid_inter} shows some representative results of methods that output detections or density maps. GeCo and CountGD locate objects well but confuse similar types, leading to overcounts. DAVE sometimes separates categories but is prone to background false positives or double-counting fragmented objects. LLMDet, despite its general vision capabilities, underperforms—likely due to limited training on dense scenes. LOCA consistently overcounts, and FamNet fails to isolate relevant objects altogether.

\subsection{Performance on inter- and intra- scenes}


In addition to using the MAE metric, we also use \emph{Normalized Absolute Error (NAE)}, a scale-aware version of MAE that adjusts for the number of objects in each image: $\text{NAE} = \frac{1}{N} \sum_{i=1}^{N} \frac{|\hat{y}_i - y_i|}{y_i}$. This enables fairer comparison across scenes with varying object counts. As shown in Table~\ref{tab:category_results}, we compare NAE between inter-class scenes (objects from different categories) and intra-class scenes (objects from the same or visually similar categories). Relative changes are reported as percentage increases or decreases when moving from inter-class to intra-class settings. Among learned models, DAVE and CountGD~(Text) are the most robust, with NAE increases of only 2.3\% and 18.1\%. LOCA surprisingly improves in intra-class settings, reducing its NAE by 19.3\%. By contrast, CountGD and GeCo show large NAE increases (+34.6\% and +20.3\%) in intra-class settings, indicating higher distractor sensitivity. FamNet and LLMDet also degrade (+5.6\% and +60.3\%).



VLMs show relatively stable NAE but high absolute errors. This stems from a tendency to default to mid- to high-range guesses (e.g., 36, 63, 67), regardless of the true count. For instance, Qwen2.5-VL often predicts counts between 30--100. Though incorrect, these guesses yield lower NAE in high-count scenes due to normalization. This behavior suggests that vision--language models may rely more on priors or prompt biases than genuine enumeration.

To understand this further, we experimented with several prompting strategies, including negation prompts that explicitly instructed models to count only one object type and \emph{not} the other. We also applied prompt engineering with more detailed descriptions and provided bounding box coordinates as reference. However, none of these techniques improved performance; in fact, they often degraded accuracy or led to further instability in count predictions.

\begin{table}[t]
  \centering
  \resizebox{\columnwidth}{!}{%
  \begin{tabular}{lrrrr}
  \toprule
  Model
    & \multicolumn{2}{c}{MAE $\downarrow$}
    & \multicolumn{2}{c}{NAE $\downarrow$} \\
  \cmidrule(lr){2-3}\cmidrule(lr){4-5}
    & Inter & Intra & Inter & Intra \\
  \midrule
  Mean Baseline           &  39.42 &  66.71 &  0.714 &  0.535 \\
  Median Baseline         &  37.38 &  57.25 &  1.587 &  0.776 \\
  \midrule
  DAVE \cite{pelhan2024dave}
                          &  46.27 &  46.75 &  0.779 &  0.797 \\
  GeCo \cite{pelhan2024novel}
                          &  45.05 &  54.80 &  0.777 &  0.935 \\
  LoCA \cite{djukic2023loca}
                          &  71.89 &  57.45 &  1.177 &  0.950 \\
  FamNet \cite{ranjan2021learning}
                          &  66.97 &  74.75 &  1.363 &  1.440 \\
  \midrule
  Count~GD \cite{amini2024countgd}
                          &  39.78 &  56.54 &  0.673 &  0.906 \\
  Count~GD~(Text) \cite{amini2024countgd}
                          &  50.23 &  53.93 &  0.712 &  0.841 \\
  LLMDet \cite{fu2025llmdet}
                          &  78.72 & 142.08 &  0.661 &  1.060 \\
  \midrule
  Ovis2 \cite{lu2024ovis}
                          &  56.87 &  74.24 &  0.711 &  0.736 \\
  Qwen2.5-VL \cite{Qwen2.5-VL}
                          &  46.35 &  67.86 &  0.598 &  0.712 \\
  LLaMA-3.2 \cite{dubey2024llama}
                          &  49.14 &  58.73 &  0.730 &  0.740 \\
  InternVL3
  \cite{zhu2025internvl3exploringadvancedtraining}
                          &  55.89 &  71.47 &  0.667 &  0.721 \\
  \bottomrule
  \end{tabular}%
  }
  \caption{MAE and NAE across inter- and intra-scenes (lower is better).}
  \label{tab:category_results}
  \end{table}

\subsection{Most and least sensitive visual attributes}

\begin{table*}[t]
\centering
\begin{tabular}{l*{3}{ccc}}
\toprule
& \multicolumn{9}{c}{\bf Evaluation on intra-category subsets (objects differ by)} \\
\multirow{2}{*}{Model} 
  & \multicolumn{3}{c}{\textbf{Color}} 
  & \multicolumn{3}{c}{\textbf{Size}} 
  & \multicolumn{3}{c}{\textbf{Texture/Shape}} \\
\cmidrule(lr){2-4} \cmidrule(lr){5-7} \cmidrule(lr){8-10}
 & MAE $\downarrow$ & RMSE $\downarrow$ & NAE $\downarrow$ 
 & MAE $\downarrow$ & RMSE $\downarrow$ & NAE $\downarrow$ 
 & MAE $\downarrow$ & RMSE $\downarrow$ & NAE $\downarrow$ \\
\midrule
Mean Baseline               & 55.16 & 83.51 & 1.698 
                            & 25.81 & 36.14 & \textbf{1.154} 
                            & 43.75 & 59.38 & 0.838 \\
Median Baseline             & 49.37 & 88.01 & 0.967 
                            & 24.71 & 36.75 & \textbf{0.586} 
                            & 40.89 & 61.71 & 0.720 \\
\midrule
DAVE \cite{pelhan2024dave}  & 63.44 & 89.16 & \textbf{0.738} 
                            & 33.26 & 39.31 & 1.293 
                            & 34.14 & 43.00 & 0.693 \\
GeCo \cite{pelhan2024novel} & 63.40 & 88.77 & \textbf{0.791} 
                            & 35.06 & 41.13 & 1.345 
                            & 52.53 & 74.55 & 0.946 \\
LoCA \cite{djukic2023loca}  & 65.37 & 95.24 & \textbf{0.799} 
                            & 33.34 & 39.66 & 1.244 
                            & 57.33 & 91.11 & 1.007 \\
FamNet \cite{ranjan2021learning}
                            & 84.92 & 117.33 & \textbf{1.296} 
                            & 56.32 & 75.44 & 1.859 
                            & 70.45 & 90.64 & 1.448 \\
\midrule
CountGD \cite{amini2024countgd} 
                            & 75.40 & 117.32 & \textbf{0.856} 
                            & 36.30 & 42.35 & 1.402 
                            & 43.95 & 57.42 & 0.793 \\
CountGD (Text) \cite{amini2024countgd}
                            & 64.51 & 98.99 & \textbf{0.760} 
                            & 38.95 & 45.61 & 1.410 
                            & 48.06 & 73.20 & 0.735 \\
LLMDet \cite{fu2025llmdet}  & 118.29 & 151.01 & \textbf{2.12} 
                            & 68.68 & 82.73 & 4.18 
                            & 89.33 & 116.06 & 2.04 \\
\bottomrule
\end{tabular}
\vskip 0.05in
\caption{Performance comparison on intra-category subsets (color, size, texture/shape).}
\label{tab:attributes_comparison}
\end{table*}

\Tref{tab:attributes_comparison} reports model performance on three subsets of the intra-category test images, divided based on how the two object subcategories in each image differ—specifically by color, size, or texture/shape. Each subset contains images where the objects differ only by the corresponding attribute. We report MAE, RMSE, and NAE, but NAE is preferred for cross-subset comparison as it normalizes object counts, making the comparison more reliable.


\textbf{Color-based variation} produces the lowest NAEs overall. DAVE (0.738) and CountGD-Text (0.760) outperform their own results on size and texture, suggesting hue is a more reliable cue than expected. GeCo also achieves low error (0.791), reinforcing that color is the easiest attribute for models to exploit.  

\textbf{Size variation}, by contrast, yields consistently higher NAEs (1.24–1.41). This challenges the intuition that spatial differences aid separation: scale changes complicate localization, particularly when small objects cluster or large ones occlude others. GeCo’s 1.345 again indicates reliance on global rather than local features.  

\textbf{Texture/shape variation} falls in between. CountGD performs best (0.735), while GeCo (0.946) and FamNet (1.448) are less robust, likely due to clutter and background interference.  

In sum, models leverage color most effectively, while size and texture distinctions remain challenging—paralleling human difficulty with such attributes. These findings underline the need for stronger feature disentanglement and attribute-specific supervision in fine-grained counting models.

\section{Conclusions}

This study presents a targeted, cross-paradigm evaluation of fine-grained counting using a diagnostic benchmark, highlighting the limitations of counters, promptable detectors, and open-source mid-sized vision-language models when precision and intent sensitivity are required. As part of this effort, we introduce PairTally, a benchmark dataset containing images with two object categories, including both inter-category and intra-category pairs. PairTally is designed to test a model's ability to count selectively in visually complex scenes. Our evaluation of ten state-of-the-art models reveals that they often fail when subtle visual distinctions are necessary—such as distinguishing pill types, identifying species, or counting similar tools—where fine-grained accuracy is critical. These findings underscore the need for more capable models, as well as more suitable training data to support their development.

{\small \myheading{Acknowledgment.} This work was funded by the Australian Institute for Machine Learning (University of Adelaide) and the Centre for Augmented Reasoning, an initiative by the Department of Education, Australian Government. The authors would also like to thank Viet Bach Tran for assistance with data collection.}

{\small
\bibliographystyle{ieee}
\bibliography{shortstrings,m_pubs_autogen,egbib}
}

\end{document}